\documentclass[journal]{IEEEtran}

\usepackage{multirow}
\usepackage[table,xcdraw]{xcolor}

\ifCLASSINFOpdf
  \usepackage[pdftex]{graphicx}
\else
\fi
\begin{document}
%
\title{A Pilot Study on the Comparison of Prefrontal Cortex Activities of Robotic Therapies on Elderly with Mild Cognitive Impairment}
%
%
%

\author{King~Tai~Henry~Au-Yeung, William Wai Lam  Chan, Kwan Yin Brian Chan, Hongjie Jiang$^{*}$, 
and~Junpei~Zhong$^{*}$,~\IEEEmembership{Senior~Member,~IEEE}
\thanks{K.T. AY, W. C, K.C, and J.Z are with the Department of Rehabilitation Sciences, The Hong Kong Polytechnic University, Hong Kong.

K.T. AY, W. C, K.C are in Master in Occupational Therapy programme, The Hong Kong Polytechnic University, Hong Kong. K.T. AY is the first author of the work.}
\thanks{H.J is with School of Shien-Ming Wu Intelligent Engineering, South China University of Technology, Guangzhou China. \\ $^{*}$J. Zhong and H. Jiang are 
Corresponding authors: \\ Junpei Zhong: joni.zhong@polyu.edu.hk, Hongjie Jiang: jiang1029@scut.edu.cn}}

%
%

\markboth{IEEE Transactions on Affective Computing}%
{Shell \MakeLowercase{\textit{et al.}}: Bare Demo of IEEEtran.cls for IEEE Journals}
%



\maketitle

\begin{abstract}
Demographic shifts have led to an increase in mild cognitive impairment (MCI), and this study investigates the effects of cognitive training (CT) and reminiscence therapy (RT) conducted by humans or socially assistive robots (SARs) on prefrontal cortex activation in elderly individuals with MCI, aiming to determine the most effective therapy-modality combination for promoting cognitive function. This pilot study employs a randomized control trial (RCT) design. Additionally, the study explores the efficacy of Reminiscence Therapy (RT) in comparison to Cognitive Training (CT). Eight MCI subjects, with a mean age of 70.125 years, were randomly assigned to ``human-led'' or ``SAR-led'' groups. Utilizing Functional Near-infrared Spectroscopy (fNIRS) to measure oxy-hemoglobin concentration changes in the dorsolateral prefrontal cortex (DLPFC), the study found no significant differences in the effects of human-led and SAR-led cognitive training on DLPFC activation. However, distinct patterns emerged in memory encoding and retrieval phases between RT and CT, shedding light on the impacts of these interventions on brain activation in the context of MCI.
\end{abstract}


%
\IEEEpeerreviewmaketitle

\section{Introduction}
%
%
%
%
Demographic shifts are occurring in some countries due to the aging of populations, leading to a notable increase in mild cognitive impairment (MCI) and its progression to dementia. The MCI manifests as a gradual decline in cognitive abilities and dementia, encompassing complex attention, executive function, learning and memory, language, perceptual-motor skills, and social cognition. The Diagnostic and Statistical Manual of Mental Disorders, Fifth Edition (DSM-5) \cite{american2013diagnostic}, introduces a unique category for social cognition within neurological cognitive illness (NCD), impacting interpersonal dynamics and elevating the risk of dementia through social isolation (National Academies of Sciences).

It has been shown that individuals with MCI can benefit significantly from social interaction \cite{smith2021loneliness,desai2020living,mahalingam2023social}, fostering positive attitudes and counteracting cognitive decline. The severity of cognitive impairment is linked to the intensity of negative experiences, while positive moods associated with social engagement enhance cognitive function and reduce the likelihood of dementia development \cite{hwang2018effects}. Chronic depression, prevalent among elderly individuals with MCI, poses a risk to cognitive abilities, emphasizing the connection of mental state and cognitive health \cite{chepenik2007influence}.

Occupational therapists typically administer therapies for individuals with Mild Cognitive Impairment (MCI). One such example is cognitive-oriented therapy, which includes cognitive training (CT), cognitive stimulation (encompassing reality orientation), and cognitive rehabilitation. Recent evidence underscores the efficacy of cognitive-oriented therapy in enhancing cognitive functions among individuals with MCI \cite{li2011cognitive}. Notably, both CT and reality orientation exhibit positive cognitive effects \cite{carrion2018cognitive}. CT is postulated to confer benefits by fostering skill improvement through repetitive practice \cite{carrion2018cognitive}, emphasizing the retention and enhancement of existing cognitive abilities to enable patients to adeptly manage contemporary tasks \cite{kasl2000psychosocial}.

Alternatively, Reminiscence Therapy (RT) concentrates on offering MCI patients materials associated with their past to evoke recollections of prior experiences \cite{saragih2022effects}. The theoretical underpinning posits that RT can facilitate the retrieval of long-term memories, consequently leading to improved cognition, enhanced quality of life, and behavioral modifications \cite{liu2017commentary}. Various objects, such as old books and pictures, capable of eliciting memories related to individuals or events from the patients' past life are employed \cite{kasl2000psychosocial}. While some studies suggest that RT may enhance cognitive function in dementia patients\cite{saragih2022effects}, not all studies concur with these findings.

Despite the effectiveness of multiple interventions in MCI, they are impractical due to scarce human resources \cite{cohen2015use}.
Therefore, cost-effective technological solutions, such as tablets, gaming consoles and robots, have been adopted \cite{ge2018technology}. Socially Assistive Robots (SARs) have been developed to facilitate more comfortable interactions, providing non-verbal and emotional interactions that could motivate and engage patients in cognitive tasks\cite{andrist2015look}. The introduction of social assistive robots (SARs) offers a potential positive impact on mood, similar to interactions with real individuals. Some studies combining cognitive treatments with SARs have shown promising results in dementia \cite{pino2020social}. The NAO robot, due to its low cost and wide-ranging capabilities, has been commonly used in human-robot studies \cite{amirova202110}. Clinically, NAO-led computerized CT was found beneficial to MCI \cite{gates2019computerised}. 
 Engaging with SARs, such as NAO, can activate the prefrontal cortex, potentially uplifting the mood of seniors with MCI, akin to conversations with humans \cite{wilson2015negative}. In this context, SARs function as companions, employing both verbal and non-verbal communication and imitating human actions through embodied structures. Joint action robots, simulating genuine interaction, utilize verbal dialogue, facial expressions, and gestures to activate cognitive functions, with respondents expressing preferences for specific interaction sides, enhancing the perceived realism of engagement with robots \cite{lukasik2021role}.

 In this project, we would like to measure the effects of RT and CT in the activation of prefrontal cortex, using both human- and robot-led interventions. Recruiting MCI elderly as subjects, to the best of our knowledge, this is the first study will answer the following questions:

 \begin{itemize}
         \item What are the effects of using SARs (i.e. the NAO robot) to conduct RT to MCI, compared with human-led interventions;
    \item What are the differences of effects on prefrontal cortex activities of using CT and RT to MCI;
    \item What is the best combination of ``therapy (CT/RT) x modality (robot-/ human-led)'' in terms of its activation on prefrontal cortex activities.
 \end{itemize}

\section{Literature Review}
\subsection{Neurophysiological Deficits with MCI and the Mechanisms for Interventions}

Short-term memory, attention, and orientation are the most reliable predictors in screening for MCI \cite{gomez2021short}. They are also predictors for early-onset AD, which is associated with poor prognosis\cite{sousa2015neural}.

The physiological underpinnings of MCI and AD are the medial temporal lobe, temporal lobe glucose metabolism, and episodic memory function for orientation \cite{sousa2015neural}, and the prefrontal cortex (PFC) for short-term memory and attention\cite{ozen2007prefrontal}. Lateral PFC is critical in attention and memory\cite{wallis2019reward}, while the dorsolateral and ventrolateral prefrontal cortices (DLPFC and VLPFC) are responsible for working memory selection, monitoring and manipulation \cite{fletcher2001frontal}, as well as working memory encoding and retrieval \cite{owen2000dissociating} respectively. DLPFC and VLPFC have close functional relationships when memory recall tasks are elicited \cite{fletcher2001frontal},  \cite{moscovitch2002frontal}. Recent research found that the DLPFC-VLPFC network relationship was emotionally-mediated, suggesting that memory encoding and retrieval are influenced by mood \cite{kirk2020cognitive}. Further, recent studies have shown hypoperfusion of oxygenated blood in the DLPFC during memory tasks (e.g. word retrieval) in elders with MCI when compared to the well elderly counterparts\cite{yeung2020functional}. On the other hand, recent research states that orientation in space, time, and person are related to the default mode network (DMN) including the medial frontal cortex, thus indicating that frontal lobe lesion is associated with orientation dysfunction \cite{peer2015brain}.

Thus, it could be deduced that the frontal lobe is one of the most susceptible regions to early-stage neurocognitive deficits. The key to preventing cognitive decline may lie in frontal lobe activation.

The finding postulates an anatomic-functional deduction that any treatments with stimuli that activate the frontal lobe (hence exercising orientation, short-term memory, and attention) can theoretically prevent neurocognitive decline, regardless of RT or CT. Thus, frontal lobe activation measurements were justified to indicate treatment effectiveness. As this study focused on the memory function, DLPFC was selected as the main region of interest (ROI).

\subsection{Social Robots for MCI}

A comprehensive review of studies investigating the use of robots in therapy for individuals with Mild Cognitive Impairment (MCI) reveals diverse interventions across various countries. Notable robots include Kompai, Paro, TIAGo Iron, and NAO, each employed in distinct experimental setups.

Studies, such as those by Agrigoroaie et al. \cite{agrigoroaie2016developing} and Chen et al. \cite{chen2022can}, highlight individual and group interventions utilizing robots like Kompai and Paro in elderly care facilities and long-term care centers. Notably, the duration of interventions ranges from 2 days to 8 weeks.

Reminiscence Therapy (RT) is explored in studies like Granata et al. \cite{granata2013robot}, where robots like Kompaï engage individuals with materials evoking past memories. This approach aims to improve cognition, quality of life, and behavior.

Additionally, investigations by Park et al. \cite{park2021humanoid} and Pino \& Palestra \cite{pino2020humanoid} utilize humanoid robots, Sil-bot and NAO, in group interventions at community centers. The studies emphasize the significance of diagnostic criteria, including the Mini-Mental State Exam (MMSE), Petersen guidelines, and cognitive function assessments.

These robotic developments as well as studies with MCI in these studies contribute valuable insights into the potential of robotic interventions for cognitive therapy in MCI.



\subsection{Using Functional Near Infrared Spectroscopy for Subjective Measurement}
In prior HRI investigations involving individuals with MCI, efficacy assessment commonly relied on subjective measures such as ratings, observations, and feedback. However, objective assessments at the neurophysiological level have been lacking. To address this gap, Functional Near Infrared Spectroscopy (fNIRS) should be employed to non-invasively measure brain activity during interventions \cite{henschel2020social,canning2013functional}. The fNIRS, capable of tracking changes in oxyhemoglobin concentration as an indicator of regional brain activity, provides a real-time, direct, and cost-effective approach for evaluating the performance of CT and RT with MCI. 

In our study, the choice of fNIRS over functional assessments like the Montreal Cognitive Assessment (MoCA) was justified due to the heterogeneous etiologies of MCI, making the anti-Alzheimer's disease (AD) deterioration effects of treatments challenging to discern within functional assessments. Additionally, improvements in cognition among MCI elders may not manifest overtly in functional assessments for a relatively prolonged period.

Moreover, fNIRS has demonstrated effectiveness in detecting cognitive activation in both robot- and human-led interventions, particularly in tasks involving memory training (i.e. CT and RT) and human-robot interactions. 
Furthermore, this study seeks to explore whether CT and RT using SARs can yield comparable benefits to human-led interventions, an area that has not been previously explored. Given inconclusive findings in prior RT studies, largely attributed to methodological flaws, this research aims to refine methodologies for a more conclusive evaluation of RT's effects in comparison to CT, both led by SARs and humans.

\section{Methodology}
\subsection{Experimental Setting}

The ``HK-MoCA 5-minutes'' \cite{wong2015montreal} was used to screen whether a subject has MCI. It is of high correlation with HK-MoCA original version ($r=0.87; p<0.001$). A total of $8$ subjects were recruited by signing up voluntarily from elderly homes and public posters.
Inclusion criteria: (1)Adult aged 50 or above, (2)Mild Cognitive Impairment (MCI), (3)have lived in Hong Kong for at least 25 years, and (4) able to comprehend Cantonese. Subjects not previously diagnosed were screened by Montreal Cognitive Assessment-5 minutes (MoCA-5), and required to score between the 2nd and 16th percentile, considering age and educational level.
Exclusion criteria were: (1)medical history of frontal lobe lesions, (2)auditory defects, (3)medical diagnosis of Parkinson’s Disease, Schizophrenia, Attention Deficit Hyperactivity Disorder (ADHD), (4)have behavioral or emotional symptoms. All these conditions would have potential implications for cognitive impairments, including attention and memory span \cite{fang2020cognition,mccutcheon2023cognitive}.

After satisfying the inclusion/exclusion criteria, all qualified participants were assigned a random code. The subjects, assessment testers, and treatment experimenters were different persons, and thus were blinded to the treatment grouping. Each participant was given HK\$100 as an incentive after having completed all the procedures, to reduce attrition.

All the participants were divided into $2$ groups. The experimental groups is summarized in Tab.~\ref{tab:groups}.
To test question 1, the recruited subjects were divided into 2 groups randomly: the ``robot-led'' group and the ``human-led'' group.
To test question 2, the subjects were then randomly divided into two groups. They were given 2 types of stimuli (verbal stories) with a clear indication of time sense to simulate CT (present information) or RT (past information):
(1)	Recent stories happening in 2023 (present-time sense), which is labeled as ``present'; and the questions for recent stories focus on verbal memory recall of the story contents, simulating cognitive training (CT);
(2)	Past stories involving elements more than 20 years ago (past-time sense), which is labeled as ``past''; and the questions for past stories focus on sharing of subject's personal experiences during the past decade depicted by the stories, simulating reminiscence therapy (RT).
To counterbalance the order effect of stimuli presentation, each group was equally divided randomly into 2 subgroups, with either the sequence of administering ``CT (present) first, RT (past) second'' (Group 1a, 2a) or ``RT (past) first, CT (present) second'' (Group 1b, 2b).

\begin{figure}
    \centering
    \includegraphics[width=0.9\linewidth]{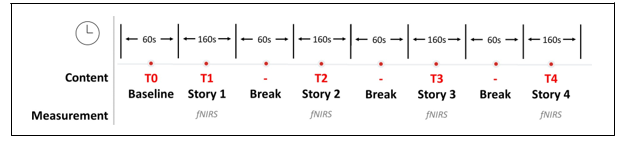}
    \caption{Study Procedure}
    \label{fig:run_down}
\end{figure}

\begin{figure}
    \centering
    \includegraphics[width=0.9\linewidth]{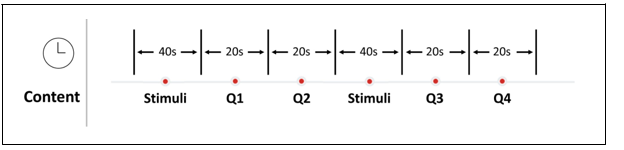}
    \centering
    \caption{Procedure of the ``Story'' session} Procedure of each story sessions (T1-T4). ``Stimuli'' refers to the verbal stimuli either given by a NAO robot or by a human. ``Q1'' to ``Q4'' are ``Q \& A'' for the participants, each lasting for $20$ seconds. 
    \label{fig:story_rundown}
\end{figure}

The experimental procedure was summarized in Figs \ref{fig:run_down} and \ref{fig:story_rundown}. One intervention session was provided to each participant. In the session, fNIRS recordings were done during the story telling (T1-T4), participants were given $1$ minute to rest to establish the signal baseline (T0). After that, each subject was given a ``Story'' interaction, Story 1 (T1), either by the NAO robot or human. Within each ``Story'' (Fig~\ref{fig:story_rundown}), participants were first given the first part of verbal stimuli (``encoding'') lasting for $40$ seconds, followed by $2$ questions (``retrieval'') lasting for a total of $40$ seconds. Then, the second part of the verbal stimulus was given for $40$ seconds and was followed by another $2$ questions. Stories $2$ to $4$ (T2-T4) were then given sequentially following the format of T1. After each interaction, participants were given a 1-minute break.

\begin{table}[]
\caption{Experimental Groups}
\label{tab:groups}
\begin{tabular}{|l|ll|ll|}
\hline
             & \multicolumn{2}{l|}{Robot-led}                                                                                             & \multicolumn{2}{l|}{Human-led}                                                                                             \\ \hline
Group Number & \multicolumn{1}{l|}{1a}                                                & 1b                                                & \multicolumn{1}{l|}{2a}                                                & 2b\\ \hline
Story 1      & \multicolumn{1}{l|}{\cellcolor[HTML]{CBCEFB}}                          &                                                   & \multicolumn{1}{l|}{\cellcolor[HTML]{CBCEFB}}                          &                                                   \\ \cline{1-1}
Story 2      & \multicolumn{1}{l|}{\multirow{-2}{*}{\cellcolor[HTML]{CBCEFB}Present}} & \multirow{-2}{*}{Past}                            & \multicolumn{1}{l|}{\multirow{-2}{*}{\cellcolor[HTML]{CBCEFB}Present}} & \multirow{-2}{*}{Past}                            \\ \hline
Story 3      & \multicolumn{1}{l|}{}                                                  & \cellcolor[HTML]{CBCEFB}                          & \multicolumn{1}{l|}{}                                                  & \cellcolor[HTML]{CBCEFB}                          \\ \cline{1-1}
Story 4      & \multicolumn{1}{l|}{\multirow{-2}{*}{Past}}                            & \multirow{-2}{*}{\cellcolor[HTML]{CBCEFB}Present} & \multicolumn{1}{l|}{\multirow{-2}{*}{Past}}                            & \multirow{-2}{*}{\cellcolor[HTML]{CBCEFB}Present} \\ \hline
\end{tabular}
\end{table}

\subsection{Experimental stimuli standardization}

The stimuli employed in CT and RT necessitate standardization due to variations in their respective protocols. In alignment with the recommendation proposed by Cotelli et al. \cite{cotelli2012reminiscence}, we implemented a standardized protocol, utilizing nearly identical storylines for CT and RT interventions. This approach aimed to eliminate potential confounding factors that might obscure the discernment of CT/RT's beneficial effects.

Our focus centered on introducing a singular distinction among stimuli, specifically the temporal orientation towards either the present or past. This deliberate manipulation sought to simulate the differential approaches of CT and RT for the purpose of standardization.

The experimental stimuli comprised succinct verbal narratives in Cantonese, addressing commonplace aspects of life and familiar objects prevalent in Hong Kong. Two distinct topics, namely ``Leisure'' and ``Living'', were chosen. Standardization ensured that each stimulus consisted of $132$ words, delivered in a neutral Cantonese tone, and maintained an equal duration for a fair comparative analysis. The language used in the narratives was drawn from Lexical Lists for Chinese Learning in Hong Kong, ensuring the comprehensibility of stimuli for participants with potentially lower educational levels. This meticulous approach ensured that the stimuli remained accessible and understandable to all study participants.

\subsection{Outcome measures}

Outcome measures were assessed starting from set up (T0) until the end of the experiment (T4).

\subsubsection{fNIRS Measurement and Removal of Confounding Noises}

We utilized a portable fNIRS device, NIRSIT (OBELAB, South Korea), specialized in measuring hemodynamic responses in the prefrontal cortex \cite{yu2020prefrontal}. The device is equipped with 32 detection sensors sampling at a rate of 8.138 Hz, generating a maximum of 204 channels.

\begin{figure}
    \centering
    \includegraphics[width=0.7\linewidth]{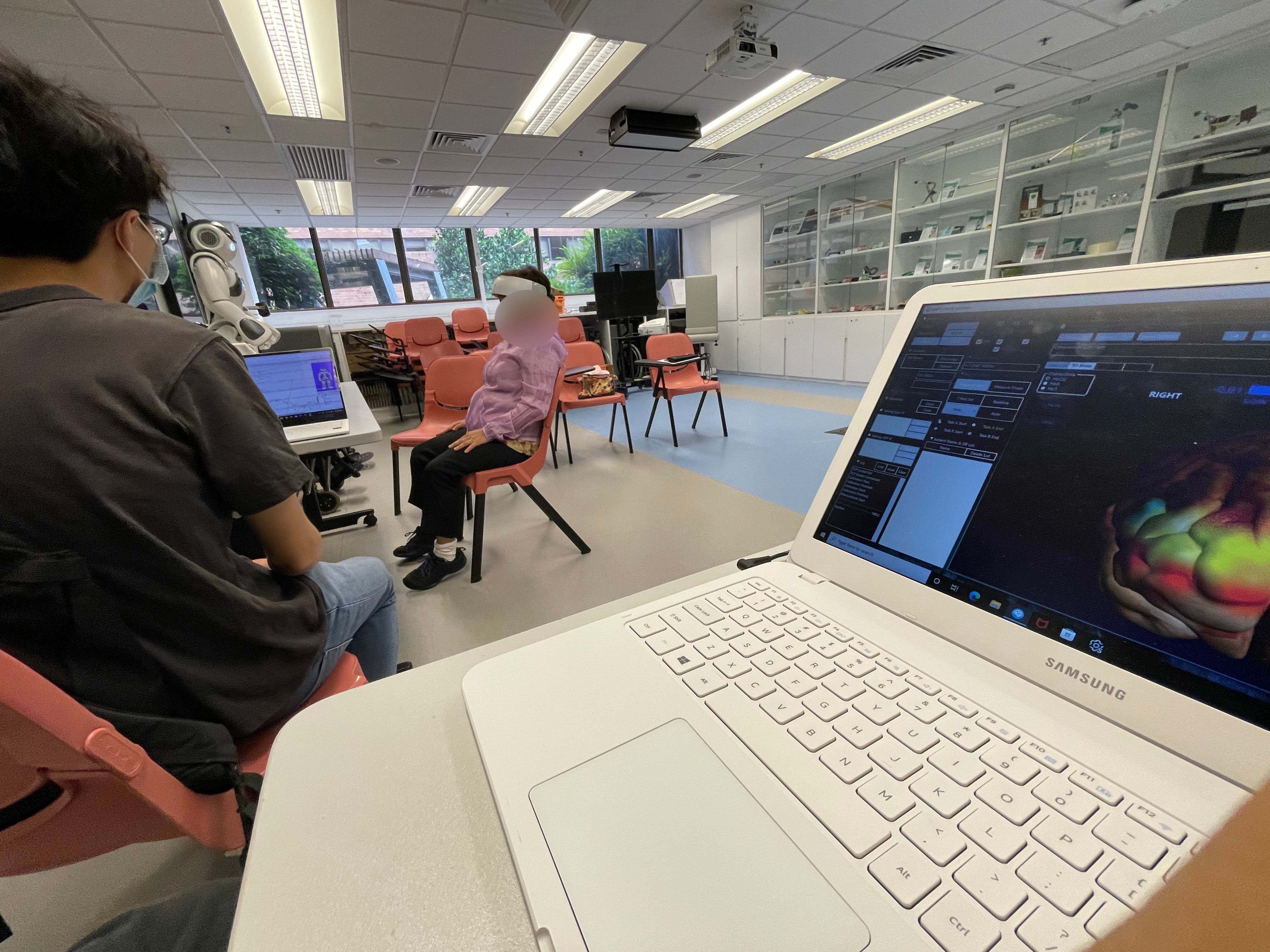}
    \caption{Robot-led Intervention with the fNIRS Measurement}
    \label{fig:fnirs-robot}
\end{figure}

To enhance data quality, several measures were implemented to minimize confounding signals or "noise." Following best-practice recommendations by Y\"{u}cel et al. \cite{yucel2021best}, the following steps were taken: (1) setting the Signal-to-Noise Ratio (SNR) threshold to 20 or above to eliminate cardiac power noises, (2) mitigating motion artifacts by instructing subjects to minimize movements and speak only when necessary, and (3) employing the "ABC-A" experimental paradigm to balance out confounding systemic signals, such as high blood pressure. Subjects were placed in rooms with minimal distraction to further ensure data accuracy. Brain activities in the frontal lobe were subsequently measured using the established fNIRS setup. The overall setup of the Robot-led intervention was shown in Fig. \ref{fig:fnirs-robot}.

\subsubsection{Acceptability of SAR}

Additionally, the Robotic Social Attributes Scale (RoSAS) developed by Carpinella et al.\cite{carpinella2017robotic} was used to measure the perception of NAO by the participants. The RoSAS has been validated empirically and uses 18 items to measure the social acceptability of robots regardless of the context and robot type.

\subsection{Data Analysis}

\subsubsection{Data Processing}

\begin{figure*}
    \centering
    \includegraphics[width=0.5\linewidth]{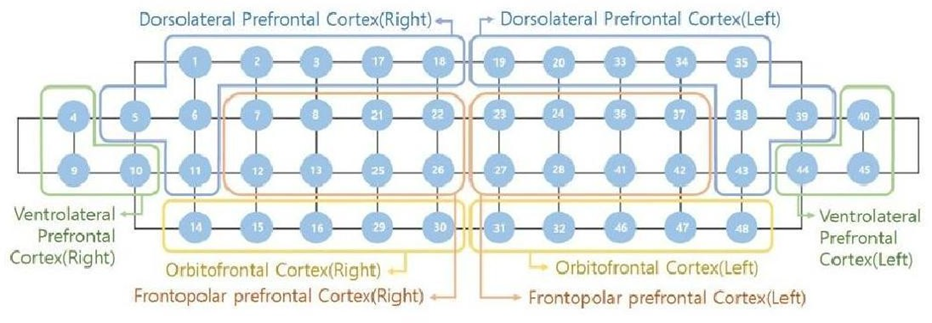}
    \caption{Grouping of the NIRSIT Channels based on the Brodmann Areas}
    \label{fig:channel}
\end{figure*}

The fNIRS data was processed using the NIRSIT analysis tool (v2.1, OBELAB). The raw data was converted to oxyhemoglobin (HbO2) and deoxyhemoglobin (HbR) concentration changes by using the modified Beer-Lambert Law \cite{hiraoka1993monte}. A cut-off frequency of $0.005 -0.1$ Hz was applied to eliminate ambient and physiological noises \cite{yoo2019diagnosis}. To minimize signal drifting, the average value from the 10s period before each task was used as the baseline for the block average. Channel signals were averaged based on the corresponding brain areas for each subject \ref{fig:channel}. The averaged data for each individual were categorized into ``encoding'' phases (each lasts for 40 seconds) and ``retrieval'' phases (each lasts for 40 seconds) based on the tasks given as Fig.\ref{fig:story_rundown} showed. These data were further grouped into ``robot-led'' group, ``human-led'' group, ``CT'' segment and ``RT'' segment for comparison.

\subsubsection{Statistical Analysis}

Baseline characteristics and channel data were compared using independent-samples t-test, paired t-test, and chi-square test. A probability of $p<0.05$ was considered statistically significant. For research question 1, independent-samples t-tests were performed: (1) between the “robot” and “human” groups, (2) between “robot-CT” and “human-CT” groups, (3) between “robot-RT” and “human-RT” groups. For research question 2, paired t-tests are performed between: (1) “CT” and “RT” interventions within subjects, (2) “CT-robot” and “RT-robot” interventions within subjects, and (3) “CT-human” and “RT-human” intervention within subjects. In each comparison, data was also finely analyzed in two different phases: ``encoding'' and ``retrieval'' phases. Statistical analyses were performed using SPSS.

\section{Results}

\subsection{Baseline}

Baseline characteristics of the subjects are listed in Tab.\ref{tab:baseline}. No significant difference was found in terms of age ($t(6)=0.068, p>0.05$), gender ($\chi^2(1)=0.533, p>0.05$), and MoCA-5-minutes baseline scores ($t(6)=-0.208, p>0.05$) between “robot-led” and “human-led” groups. There was no attrition in the study.

\begin{table*}[]
\centering
\caption{Baseline Characteristics of Subjects}
\label{tab:baseline}
\begin{tabular}{|l|l|l|l|l|}
\hline
              & Robot-led        & Human-led        & $t(6)$ or $\chi^2(1)$ & $p$     \\ \hline
Age           & $70.2500\pm6.02080$ & $70.0000\pm4.24264$ & $0.08 $         & $0.948$ \\ \hline
Gender (Male) & $2 (50\%) $        & $1 (25\%) $        & $0.533 $        & $0.465$ \\ \hline
MoCA Score    & $19.0000\pm2.61406$ & $19.3750\pm2.49583$ & $-0.208 $       &$ 0.842$ \\ \hline
\end{tabular}
\end{table*}

\subsection{Between-group comparison between ``robot-'' and ``human-led'' interventions}

HbO2 changes from the baseline of the DLPFC in ``robot-'' and ``human-led'' groups were shown in Tab.\ref{tab:dlpfc}. During the encoding and retrieval phase, there was no significant difference between ``robot-led'' and ``human-led” groups (encoding phase: right: $t(6)=1.284, p=0.247$; left: $t(6)=1.341, p=0.229$)(retrieval phase: right: $t(6)=0.017, p=0.987$; left : $t(6)=-0.206, p=0.843$). Both groups gave negative HbO2 changes across left and right cortices during the encoding phase, while in the retrieval phase, both groups gave positive HbO2 changes.

\begin{table*}
	\centering
	\caption{Comparison of HbO2 changes in the DLPFC during robot- and human-led Intervention}
	\label{tab:dlpfc}
	\begin{tabular}{|p{1.5cm}|p{2cm}|p{6cm}|p{6cm}|}
		\hline
		Comparison & Condition & dPFC, right & dPFC, left \\ \hline
		\multirow{6}{*}{Robot- \& Human-led} & Encoding phase (all inclusive) & $NAO=-0.00000925\pm0.00031734; Human=-0.00028580\pm0.000291340, t=1.284, p=0.247$ & $NAO=-0.00008150\pm0.00017997; Human=-0.00038297\pm0.00041221, t=1.341, p=0.229$ \\ \cline{2-4}
		& Retrieval phase (all inclusive) & $NAO=0.00008987\pm0.00015995; Human=0.00008840\pm0.00007983, t=0.017, p=0.987$ & $NAO=-0.00008227\pm0.00011394; Human=0.00011452\pm0.000291100, t=-0.206, p=0.843$ \\ \cline{2-4}
		& CT, encoding phase & $NAO=-0.00005543\pm0.00025933; Human=-0.00023994\pm0.00037567, t=0.808, p=0.450$ & $NAO=-0.00004328\pm0.00022114; Human=0.00000276\pm0.00039315, t=-0.204, p=0.845$ \\ \cline{2-4}
		& RT, encoding phase & $NAO=0.00003687\pm0.00037755; Human=-0.00033185\pm0.00022452, t=1.679, p=0.144$ & $NAO=-0.00011968\pm0.00024224; Human=-0.00076874\pm0.00046551, t=2.474, p=0.048^*$ \\ \cline{2-4}
		& CT, retrieval phase & $NAO=0.00006849\pm0.00021184; Human=0.00001517\pm0.00030186, t=0.289, p=0.782$ & $NAO=0.00002122\pm0.00016662; Human=-0.00005532\pm0.00037510, t=0.373, p=0.722$ \\ \cline{2-4}
		& RT, retrieval phase & $NAO=0.00011141\pm0.00015298; Human=0.00016193\pm0.00017686, t=-0.432, p=0.681$ & $NAO=0.00014311\pm0.00006981; Human=0.00028435\pm0.00028277, t=-0.970, p=0.370$ \\ \hline
	\end{tabular}
\end{table*}

\subsection{Comparison between HbO2 changes in CT and RT}

HbO2 changes from baseline in the DLPFC during CT and RT stories were compared using paired t-test in Tab.\ref{tab:robot-human}. During both encoding and retrieval phases, a significant difference was found between CT and RT in the left DLPFC (Encoding: $t(7)=2.668, p=0.032$;
Retrieval: $t(7)=-2.592, p=0.036$). In the encoding phase, both CT and RT gave negative HbO2 values, with RT giving more negative HbO2 change. During the retrieval phase, RT gave positive HbO2 changes, in contrast to the negative change in CT.
Comparing CT and RT with the consideration of intervention medium (robot or
human), the only statistically significant difference was found in human-led group x
encoding phase in left DLPFC, where CT gave significantly higher HbO2 change (though close to zero) than the negative change in RT ($t(7)=6.146, p=0.009$).

\begin{table*}[]
\centering
\caption{Comparison of subject HbO2 changes in the DLPFC during CT and RT}
\label{tab:robot-human}
\begin{tabular}{|p{1.5cm}|p{1cm}|p{5cm}|p{5cm}|}
\hline
Comparison                   & Condition       & dPFC, right                                                              & dPFC, left                                                               \\ \hline
\multirow{2}{*}{Both groups} & Encoding phase  & $CT=-0.00014768\pm0.00031469, RT=-0.00014748\pm0.00034863,t=-0.003,p=0.998$  & $CT=-0.00002025\pm0.00029632, RT=-0.00044421\pm0.00048825,t=2.668,p=0.032*$  \\ \cline{2-4} 
                             & Retrieval phase & $CT=-0.00004183\pm0.000243093, RT=-0.00013667\pm0.00015545,t=-0.803,p=0.448$ & $CT=-0.00001705\pm0.00027179, RT=0.00021373\pm0.00020508,t=-2.592,p=0.036*$  \\ \hline
\multirow{2}{*}{Robot-led}   & Encoding phase  & $CT=-0.00005543\pm0.00025933, RT=-0.00003687\pm0.00037755,t=-1.397,p=0.258$  & $CT=-0.00004328\pm0.00022114, RT=-0.00011968\pm0.00024224,t=-0.521,p=0.638$  \\ \cline{2-4} 
                             & Retrieval phase & $CT=0.00006849\pm0.00021184, RT=0.000111417\pm0.00015298,t=-0.465,p=0.673$   & $CT=0.00002122\pm0.00016662, RT=0.00014311\pm0.00006981,t=-2.097,p=0.127$    \\ \hline
\multirow{2}{*}{Human-led}   & Encoding phase  & $CT=-0.00023994\pm0.00037567, RT=-0.00033185\pm0.00022452,t=0.884,p=0.442$   & $CT=-0.00000276\pm0.00039315, RT=-0.00076874\pm0.00046551,t=6.146,p=0.009**$ \\ \cline{2-4} 
                             & Retrieval phase & $CT=0.00001517\pm0.00030186, RT=0.00016193\pm0.00017686,t=-0.627,p=0.575 $   &$ CT=-0.00005532\pm0.00037510, RT=0.00028435\pm0.00028277,t=-2.118,p=0.124 $  \\ \hline
\end{tabular}
\end{table*}

\subsection{Acceptability of SAR}

\begin{figure*}
    \centering
    \includegraphics[width=0.7\linewidth]{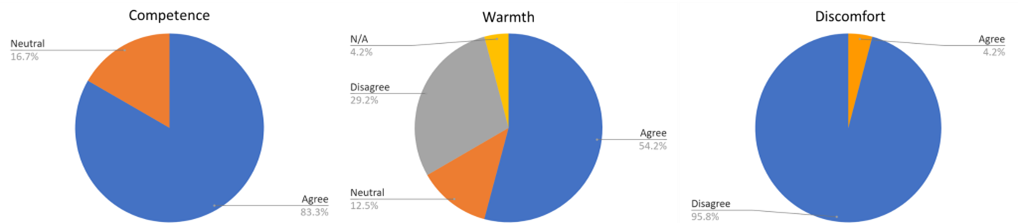}
    \caption{Summary of the RoSAS}
    \label{fig:rosas}
\end{figure*}

Using the SAR survey, most subjects from the robot-led group expressed positive opinions towards the NAO robot (83.3\% agree with ``competence'', 54.2\% agree with ``warmth''). Only a small proportion of subjects expressed negative sentiment towards the SAR (4.2\% agree with ``discomfort'') (Fig. \ref{fig:rosas}.

\section{Discussion}

\subsection{Comparative Analysis of the Impact of Human- and Robot-Led Interventions on MCI}

The alterations in HbO2 concentration within the DLPFC during both robot-led and human-led interventions did not exhibit significant differences. This suggests a similarity in DLPFC activation between the two intervention modalities, indicating that robot-led interventions may evoke brain activation patterns comparable to those observed in traditional human-led interventions. Recent findings supporting the positive impact of NAO robot-led cognitive training further underscore this notion \cite{amirova202110}. This study pioneers the comparison of such effects with therapies conducted by human practitioners, suggesting that robot-led interventions could serve as a viable alternative in therapy sessions for elders with MCI.

In the context of the memory retrieval phase, both robot-led and human-led groups displayed increased DLPFC activation, as evidenced by positive HbO2 changes. This aligns with contemporary research highlighting PFC activation in MCI subjects during memory retrieval phase \cite{heun2007mild, petrella2006mild, uemura2016reduced}, albeit some results of these studies showed the PFC activation levels in MCI during memory retrieval phase were actually lower than those of the cognitively-healthy elders. Consequently, memory retrieval tasks mediated by both humans and robots may induce similar PFC activation patterns.

Contrastingly, during the memory encoding phase in both robot and human groups, the DLPFC exhibited reduced activation, i.e. deactivation, compared to the resting state. This contradicts established studies suggesting PFC activation during memory encoding tasks \cite{petrella2006mild, uemura2016reduced}. However, some studies have reported PFC hypoperfusion in MCI patients during memory tasks, where the PFC is less activated compared to those with normal cognitive function \cite{yeung2020functional}. It has also been observed that lateral prefrontal and parietal brain regions are activated during tasks involving memory and executive function \cite{yeung2020functional, bokde2010altered}. Bokde et al. \cite{bokde2010altered} proposed that this phenomenon could be attributed to limited brain resources in MCI, leading to an impaired ability to compensate for age-associated hypoperfusion in the PFC. Hence, a similar explanation could be posited for the reduced DLPFC activation in this study: during memory encoding tasks, brain regions other than the PFC might have been activated, diverting oxygenated blood supply away from the PFC. Considering the limited pool of brain resources in elders with MCI, this might result in diminished oxygenation in the PFC.

\section{Comparative Analysis: Reminiscence Therapy (RT) vs. Cognitive Training (CT)}

\subsection{Encoding Phase}

As two of the most widely utilized therapeutic approaches for Mild Cognitive Impairment (MCI), RT and CT exhibit three significant differences during the encoding phase.

The first crucial distinction is evident in the HbO2 concentration changes in the left DLPFC. Both RT and CT display negative values during this phase, signifying deactivation. However, RT results in more negative values than CT, indicating a greater deactivation of the left DLPFC when processing past-information in RT compared to present-information in CT.

Our study pioneers the investigation of the therapeutic effects of RT on MCI elders using fNIRS brain activation data. Unlike prior research that primarily focused on cognitively intact elders' memory retrieval phase, our finer analysis delves into both memory encoding and retrieval phases. Notably, studies before 2023 relied on post-intervention outcome measurements, such as functional, cognitive, or mood assessments, yielding inconclusive results \cite{woods2018reminiscence}.

One plausible explanation for DLPFC deactivation during verbal memory encoding aligns with Bokde et al.'s postulation that MCI elders experience age-related hypoperfusion in the PFC \cite{bokde2010altered}. The cognitive demands of the encoding process, perceived as cognitively demanding by the subjects, may lead to DLPFC hypoperfusion as limited cognitive resources are directed to other brain regions.

These findings find support in other fNIRS/fMRI studies exploring verbal memory encoding and neuro-correlates. Emch et al. \cite{emch2019neural} revealed that high attentional control in verbal memory encoding led to holistic cerebral blood flow distribution to the PFC, hippocampus, parietal cortex, and cerebellum. In elders with MCI, facing cognitively demanding verbal memory encoding triggered blood flow to the entire PFC and other structures, outside the left DLPFC, explaining its hypoperfusion.

Further, RT encoding caused even more hypoperfusion than CT encoding, and previous fNIRS/fMRI studies provide insights:

\begin{enumerate}
  \item Carelli \& Olsson \cite{carelli2014neural} demonstrated that present time-sense activates superior medial frontal cortex (BA32), while past time-sense activates medial frontal cortex (BA10). As the left DLPFC is located in BA33-35, closer to present-time activated cortex, CT induces less hypoperfusion compared to RT, where blood flow must travel a longer distance to activate BA10.
  
  \item Skye et al. \cite{skye2023localization} emphasized the involvement of the time-orientation network (precuneus, medial temporal lobes, and occipitotemporal cortex) in time orientation, integral for MCI elders. RT's longer time-distance than CT in the universal time frame drains more blood to the whole time-orientation network from the left DLPFC, causing increased hypoperfusion.
  
  \item Ferreri et al. \cite{ferreri2013music,ferreri2014less} attributed older adults' DLPFC reduced activation to facilitating semantic background in stimuli, for example background music-playing instead provided a richer context to lessen DLPFC verbal memory encoding demand and hence reduced activation, compared to silence. In our study, RT's story content and obsolete lifestyles related to the "1970s" are highly time-specific and contrast distinctively to subjects' present-day lives, compared to CT's less time-specific modern lifestyles as described in "2023". Therefore, "1970s" time-specific semantic background could benefit MCIs' verbal memory encoding, by providing a helpful, highly distinctive time-context and therefore less demand on DLPFC, leading to increased hypoperfusion.
\end{enumerate}

\subsection{Memory Encoding Phase - Human-led Group}

The second significant difference emerges during the memory encoding phase in the human-led group. HbO2 concentration changes in the left DLPFC show a significant difference between CT and RT. RT results in negative values, indicating deactivation, while CT shows slightly positive values. This suggests more prominent left DLPFC deactivation when processing past-information in RT, particularly in the human-led group.

The explanation builds upon previous paragraphs, with additional considerations for the human-led group. In this context, MCI subjects, compared to the robot-led group, manage RT past-information stimuli from human speakers, demanding attention to facial features. Bardon et al. \cite{bardon2022face} noted the activation of "face neurons" in the inferior temporal cortex, further draining limited cognitive resources from the left DLPFC in MCI elders. With MCI's inherent limited attentional resources to verbal memory encoding and RT's past time-sense, the simultaneous tri-functional demand exacerbates left DLPFC hypoperfusion during RT encoding in the human-led group.

\subsection{Retrieval Phase}

The final significant difference surfaces in HbO2 concentration changes in the left DLPFC between CT and RT during the retrieval phase. RT yields positive values, indicating activation, while CT shows negative values. This suggests left DLPFC activation during sharing familiar past in RT, contrasting with deactivation during recalling recently-heard story-stimuli in front of experimenters.

The positive effect of RT during retrieval is echoed by Kobayashi et al. \cite{kobayashi2023reminiscence}, linking frontal lobe activation to positive mood induction during RT memory retrieval tasks. This emphasizes the importance of considering RT as a mood-oriented intervention, with the DLPFC-VLPFC network being emotionally-mediated in verbal memory tasks \cite{kirk2020cognitive}.

CT's brain deactivation during retrieval may be explained by cognitive demands, as discussed earlier. Notably, CT shows positive values in the robot-led group but negative values in the human-led group during the retrieval phase. This discrepancy may be attributed to elders' self-ego or anxiety issues about performance in front of humans, causing brain deactivation. Socially-assistive robots may offer advantages in CT for elders with MCI, alleviating performance anxiety \cite{tseng2018functional}.

\subsection{Limitations and Future Directions}

This study has several limitations. Firstly, the sample size is small due to practical and study design concerns. The study faced constraints related to budget limitations and time constraints in recruiting subjects. Additionally, there was a limited availability of subjects diagnosed with MCI in the community. On the positive side, the study was conducted on a pilot basis, focusing on providing preliminary data for a larger-scale study and checking the feasibility of the proposed protocol. However, the small sample size may introduce bias, and any conclusions derived from statistical analysis should be approached with caution.

Furthermore, the absence of a control group in the study means that the effectiveness of the intervention compared to non-intervention could not be determined.

Future studies should prioritize a larger sample size to enhance the reliability and generalizability of the findings. Including control groups would provide valuable insights into the comparative effectiveness of interventions for elders with MCI. The experimental protocol can be refined by employing clearer instructions to minimize confusion among subjects and adopting more flexible protocols. Measures such as using chairs with headrests can be implemented to mitigate motion artifacts.

Exploring future directions, the application of SARs in CT and RT warrants investigation. Subsequent studies could delve into the brain activation patterns during the memory encoding and retrieval phases. Given the recorded reduced activation of the PFC during memory encoding, a larger sample size in future studies can validate and elucidate the observed brain activation patterns. Moreover, as RT has demonstrated comparable effects to CT, especially in therapies involving the sharing of past experiences, future research should explore the influence of subject engagement levels and anxiety factors. This avenue presents a promising direction for further research to clarify the inconclusive effects noted in previous studies with methodological flaws. It is anticipated that future research, incorporating larger sample sizes and methodological refinements, will contribute to a more robust affirmation of the therapeutic effects of RT.

\section{Conclusion}

In conclusion, robot-led interventions demonstrate comparable effectiveness to human-led interventions in addressing cognitive challenges among elders with Mild Cognitive Impairment (MCI). The positive reception of Socially Assistive Robots (SARs) by the subjects suggests their potential as a novel therapeutic approach in MCI rehabilitation.

However, Cognitive Training (CT) and Reminiscence Therapy (RT) exhibit distinct effects on brain activation. Specifically, the act of listening to information during the memory encoding phase may lead to Prefrontal Cortex (PFC) deactivation, indicating a cognitive demand among elders with MCI. Notably, listening to past information (RT) appears to impose a greater cognitive demand compared to listening to present information (CT). This pattern is evident in the human-led group but not observed in the robot-led group. Interestingly, discussing past information during RT prompts more PFC activation than recalling present information, suggesting that RT may offer advantages over CT when encouraging elders with MCI to share their familiar past stories. The underlying mechanisms behind these findings were explored, and the study proposes several clinical implications and avenues for future research.

\section*{Acknowledgment}

The authors would like to thank Professor Sam Chan, Professor Stella Cheng, Ms. Vera Lam and Ms. Anita Ngan, The Hong Kong Polytechnic University, for their occupational therapy teaching about dementia and ageing as well as countless support throughout this study. 

Mr. Au-Yeung King Tai, the first author, would also like to thank Mr. Johnny Lam, The Hong Kong Polytechnic University, for his advice as the author's academic advisor.

\ifCLASSOPTIONcaptionsoff
  \newpage
\fi

\bibliographystyle{IEEEtran}
\bibliography{sar-paper}







\end{document}